  \documentclass[graybox]{svmult}

  \usepackage{mathptmx}       
  \usepackage{helvet}         
  \usepackage{courier}        
  \usepackage{type1cm}        
  \usepackage{makeidx}         
  \usepackage{graphicx}        
  \usepackage{multicol}        
  \usepackage[bottom]{footmisc}
  \usepackage{tikz}
  \usepackage{caption}
  \usepackage{stfloats}

  \usepackage{amsmath}

  \newcommand{\btheta}{\mathbf{\theta}}

  \newcommand{\bo}{\mathbf{o}}
  \newcommand{\ba}{\mathbf{a}}
  
  \newcommand{\bc}{\mathbf{c}}

  \newcommand{\bn}{\mathbf{n}}
  \newcommand{\bs}{\mathbf{s}}
  \newcommand{\bh}{\mathbf{h}}
  \newcommand{\bT}{\mathbf{T}}
  \newcommand{\bK}{\mathbf{K}}
  \newcommand{\bC}{\mathbf{C}}
  \newcommand{\etal}{{\em et al.}}

  \makeindex             

\begin{document}

  \title*{Learning to Singulate Objects using a Push Proposal Network}
  \author{Andreas Eitel, Nico Hauff and Wolfram Burgard}
  \institute{All authors are with the Department of Computer Science, University of Freiburg, Germany. Corresponding author's email: eitel@cs.uni-freiburg.de}
  %
  %
  \maketitle

  \abstract{Learning to act in unstructured environments, such as cluttered piles of objects, poses a substantial challenge for manipulation robots.
  We present a novel neural network-based approach that separates unknown objects in clutter by selecting favourable push actions. Our network is trained from data collected through autonomous interaction of a PR2 robot with randomly organized tabletop scenes. The model is designed to propose meaningful push actions based on over-segmented RGB-D images. We evaluate our approach by singulating up to 8 unknown objects in clutter. We demonstrate that our method enables the robot to perform the task with a high success rate and a low number of required push actions. Our results based on real-world experiments show that our network is able to generalize to novel objects of various sizes and shapes, as well as to arbitrary object configurations.
  Videos of our experiments can be viewed at ~\url{http://robotpush.cs.uni-freiburg.de}}

  \section{Introduction}
  \label{sec:1} 

  Robot manipulation tasks such as tidying up a room or sorting piles of objects are a substantial challenge for robots, especially in scenarios with unknown objects. The objective of object singulation is to separate a set of cluttered objects through manipulation, a capability regarded as relevant for service robots operating in unstructured household environments. Further, the ability to separate unknown objects from surrounding objects provides great benefit for object detection, which still remains an open problem for overlapping objects. A key challenge of object singulation is the required interaction of multiple capabilities such as perception, manipulation and motion planning. Perception may fail because of occlusions from objects but also from occlusion by the manipulator itself. Manipulation actions such as grasping may also fail when attempting to manipulate objects that the perception systems fails to correctly segment. Motion planning is particularly prone to errors when applied in unknown, unstructured environments. 

  \begin{figure}[t]
  \centering
  \includegraphics[width=0.99\columnwidth]{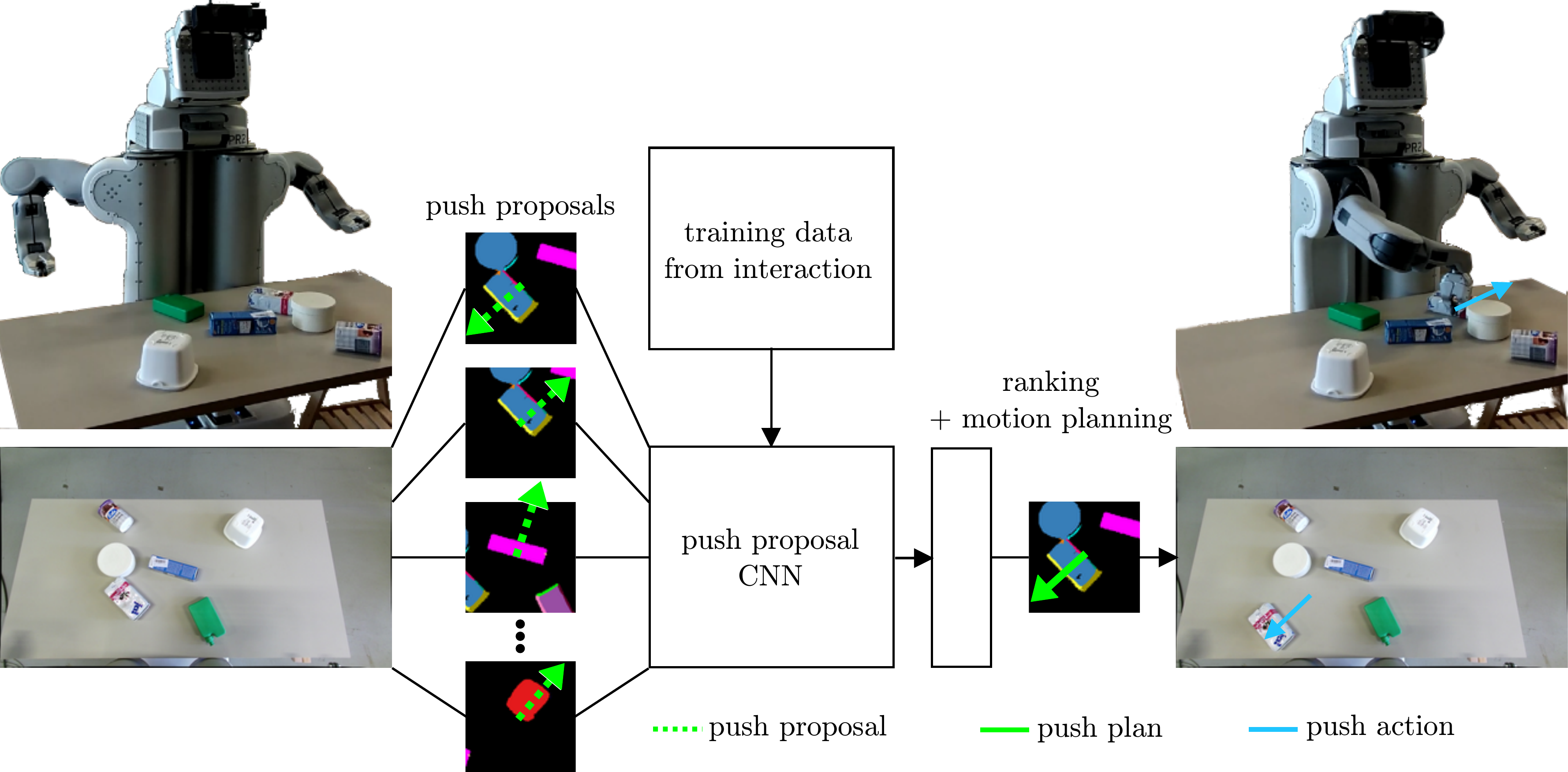}
  \caption{Our approach consists of a convolutional neural network that ranks a set of potential push actions based on an over-segmented input image in order to clear objects in clutter. The proposed pushes are executed using motion planning. Our push proposal network is trained in an iterative manner from autonomous interaction with cluttered object scenes.}
  \label{fig:approach}       
  \end{figure}

  Previous approaches for object singulation strongly incorporate the concept of object segments or even complete objects. Push action strategies are selected based on the current belief of the configuration of all segments or objects in the scene, an assumption that is not robust, since segments might merge and object detection might fail.
  We apply a different strategy that aims to relax the concept of segments and objects for solving the task at hand, making task execution less prone to modelling errors from the perception modules.
  Similar to recent approaches that operate directly on image inputs~\cite{levine2016end, mahler2017dex} we aim for an end-to-end action selection approach that takes as input an RGB-D image of a tabletop scene and proposes a set of meaningful push actions using a convolutional neural network, as shown in Fig.~\ref{fig:approach}.

  Our primary contributions are: 1) a push proposal network (Push-CNN), trained in an iterative manner to detect push candidates from RGB-D images in order to separate cluttered objects, 2) a method that samples potential push candidates based on depth data, which are ranked with our Push-CNN, and 3) real-world experiments with a PR2 robot, in which we compare the performance of our method against a strong manually-designed baseline.
  
	We quantitatively evaluate our approach on 4 sets of real-robot experiments. These experiments involve singulation tasks of increasing difficulty ranging from 4 unknown objects up to 
	configurations with 8 unknown objects. In total the robot executed over $400$ push actions during the evaluation and achieved an object separation success rate of up to $70\%$, which shows that our Push-CNN is able to generalize to previously unseen object configurations and object shapes. 

  \section{Related Work}
  \label{sec:2}
  Standard model-based approaches for object singulation require knowledge of object properties to perform motion planning in a physics simulator and to choose push actions accordingly~\cite{cosgun2011push,dogar2011framework}. However, estimating the location of objects and other properties of the physical environment can be subject to errors, especially for unknown objects~\cite{yu2016more}. Interactive model-free methods have been applied to solve the task by accumulating evidence of singulated items over a history of interactions including push and grasp primitives~\cite{chang2012interactive}. Difficulties arise when objects have to be tracked after each interaction step, requiring small motion pushes due to partial observations and occlusions. We follow a model-free approach that encodes evidence of singulated objects in the learned feature space of the network. 
  Hermans~\etal~\cite{hermans2012guided} perform object singulation using several push primitives similar to ours.
  Their method is based on object edges to detect splitting locations between potential objects to apply pushes in those regions respectively, but does not include learned features and does not take into account stacked objects.
  Katz~\etal~\cite{katz2014perceiving} present an interactive segmentation algorithm to singulate cluttered objects using pushing and grasping. Similar to them, we also create object hypothesis from over-segmenting objects into surface facets but use a different method based on the work of~\cite{richtsfeld2012segmentation}. 
  They perform supervised learning with manual features to detect good manipulation actions such as push, pull and grasp. We take the idea of learning a step further by directly learning from over-segmented images and therefore removing the need for manual feature design.
  Boularias~\etal~\cite{Boularias2015} learn favourable push actions to improve grasping of objects in clutter in a reinforcement learning setting, but use manual features and do not show the applicability of their method for more than 2 objects.     
  Gupta~\etal~\cite{gupta2015using} present an approach to sort small cluttered objects using a set of motion primitives, but do not show experiments with  objects of various sizes and shapes.
  Laskey~\etal~\cite{laskey2017comparing} leverage learning from human demonstrations
  to learn control policies for object singulation, but thus far only considered 
  singulating a single object from a cluttered scene, 
  while our method can singulate up to 6 objects and achieve a higher success rate.
  \\
  Our approach is related to robot learning from physical interaction~\cite{pinto2016curious}, including approaches that learn to predict how rigid objects behave if manipulated~\cite{kopicki2011learning,hermans2013learning,zhu2017model} and self-supervised learning methods for grasping~\cite{pinto2016supersizing,levine2016learning}.
  Gualtieri~\etal~\cite{gualtieri2016high} present a grasp pose detection method in which they detect grasp candidates using a convolutional neural network. They follow a similar methodology but for a different task and goal.
  Recent methods learn the dynamics of robot-object interaction for the task of pushing objects to a target location~\cite{finn2016deep,agrawal2016learning}, but do not evaluate on scenes with many touching objects and do not follow an active singulation strategy. Byravan~\etal~\cite{byravan2016se3} learn to predict rigid body motions of pushed objects using a neural network, but to not demonstrate results for multiple objects in a real-world scenario.      
  Object separation is used as a source to leverage perceptual information from interaction following the paradigm of interactive perception~\cite{bohg2016interactive}. As demonstrated in our experiments, our singulation method can be used to generate informative sensory signals 
  that lower the scene complexity for interactive object segmentation~\cite{schiebener2011segmentation,hausman2013tracking,van2014probabilistic, koo2014unsupervised}.

  \begin{figure}[t]
  \centering
  \includegraphics[width=0.99\columnwidth]{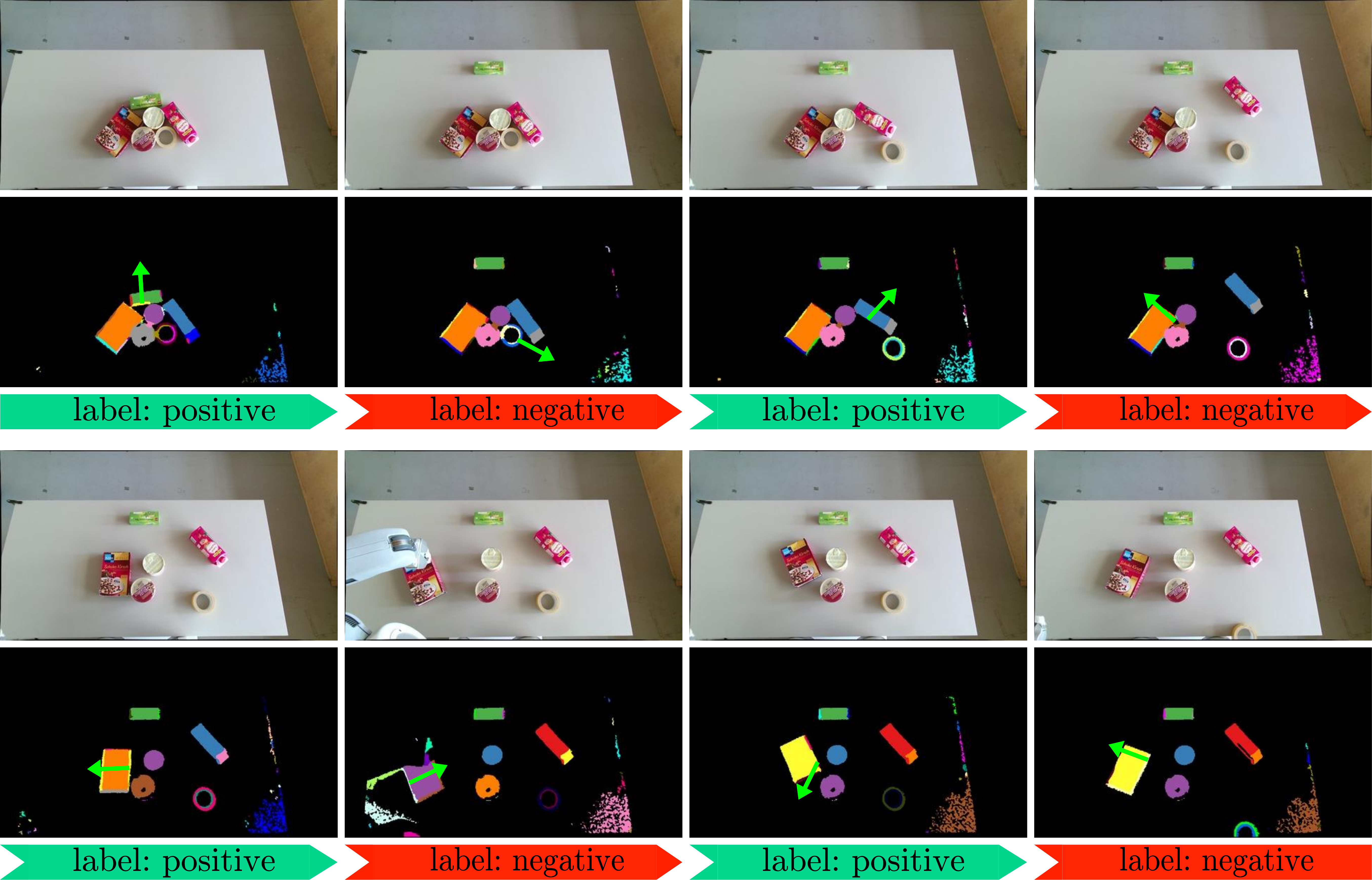}
  \caption{For training the push proposal CNN each push action (small arrow) from the collected dataset is labeled as positive or negative by a user who assesses the outcome of the action. The criteria for a positive label are: the object got singulated, the push did not move multiple objects, the object was pushed close to its center of mass and the pushed object was not already singulated.}
  \label{fig:training_data}       
  \end{figure}

  \section{Learning to Singulate Objects}
  We consider the problem of learning good push actions to singulate objects in clutter.
  Let $\bo$ be an image with height $H$ and width $W$ taken from an RGB-D camera with a known camera intrinsics matrix $\bK$. Further, let $\ba = (\bc,\alpha)$ be a push proposal with start position $\bc = (x,y)$ and push angle $\alpha$ both specified in the image plane.
  We aim to learn a function $p = F(\bo,\ba;\btheta)$ with learned parameters $\btheta$  that takes as input the image together with a push proposal and outputs a probability of singulation success.

  \subsection{Data}
  The prediction function $F$ is trained in an iterative manner (we use 2 training iterations $F_1,F_2$) on experiences that the robot collects trough interaction with the environment.    
  At the first iteration, we gather a dataset $\mathcal{D}$ of push proposals together with images from randomly pushing objects in simulation  .
  Then, we train a classifier $F_1$ that best mimics the labels from an expert user in terms of successful and unsuccessful singulation actions.
  In the next iteration, we use $F_1$ to collect more data and add this data to $\mathcal{D}$.
  The next classifier $F_2$ is trained to mimic the expert labels on the whole dataset $\mathcal{D}$.
  We refer to $F_1$ as a vanilla network that is trained solely on experience collected from random interaction and $F_2$ as an aggregated network.
  The training examples $\mathcal{D} = \lbrace (\bo^{1}, \ba^{1}, y^{1}), \dots, (\bo^{N}, \ba^{N}, y^{N}) \rbrace$ are labeled by an expert user, where $y^i$ is a corresponding binary label. Fig.~\ref{fig:training_data} depicts the labels given by an expert user for one exemplary singulation trial.
  The function is trained in a supervised manner using a negative log likelihood loss $\min_{\substack \btheta} \sum_{i=1}^{N} \mathcal{L}\left (p, y^i \right)$ and a sigmoid function.

  \subsection{Approach}
  Our approach is divided into three modules, as shown in Fig.~\ref{fig:approach}: 1) a sampling module that generates a set of push proposals, 2) a neural network module that classifies the set of  push proposals and ranks them accordingly, and 3)  an action execution module that computes arm motion plans for the ranked pushes and then executes the first push proposal that has both a high probability of singulation success and a successful motion plan.

  \subsubsection{Push Proposal Sampling}
  \label{Sampling}
  Our push proposal sampling method is designed to generate a set of push proposals $\{ \ba^1,\dots, \ba^M \}$.
  First, we sample 3D push handles $\{ \bh^1,\dots, \bh^M \}$ represented as point normals pointing parallel to the xy-plane of the robot's odometry frame $\bh^m=(x,y,z,\bn_h)$, $\bn_h=(-n_x,-n_y,0)$. Specifically, given the raw depth image, we first apply a surface-based segmentation method~\cite{richtsfeld2012segmentation} (we only use the pre-segmentation and surface-detection steps of the approach) to obtain a set of segments 
  $\{ \bs^1,\dots, \bs^L \}$ together with a surface normals map and a table plane.
  Second, we sample for each segment $\bs^l$ a fixed number of push handles and remove push handles below the table plane. We assume a fixed push length $l_a=0.2$ for all handles and compute the convex hull for the table in order 
  to filter out push handles that have push end-points outside of the table.
  Finally, we obtain a set of push proposals by transforming the push handles into the camera frame using the transformation matrix $\bC$ from the odometry to the camera frame together with the camera intrinsics $\ba=\bK\bC\bh$.

  \subsubsection{Push Proposal Network and Input}

\begin{figure}[t]
   \centering
   \includegraphics[width=0.90\columnwidth]{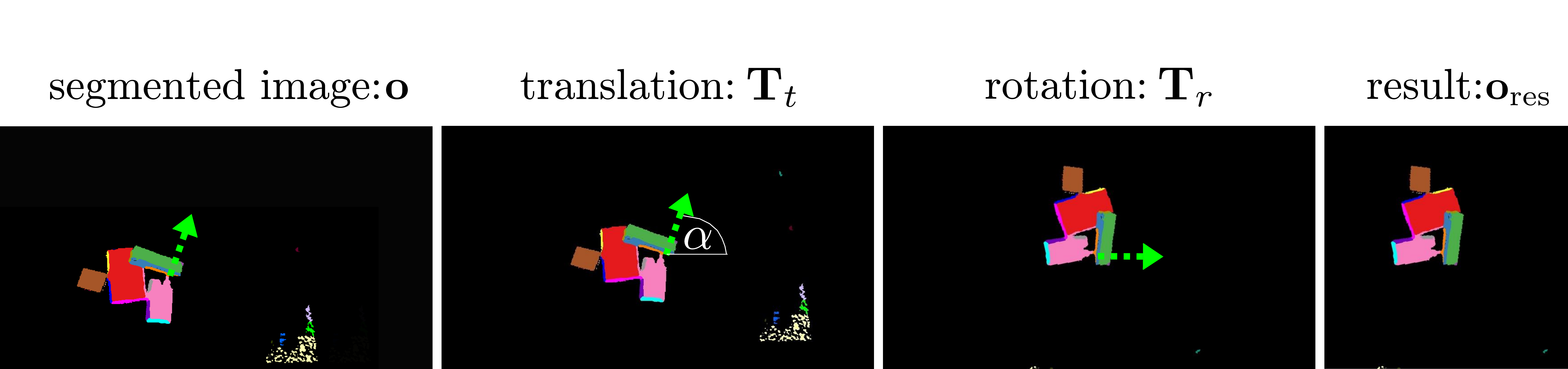}
   \caption{We encode $\bo$ and $\ba$ together in a push proposal image $\bo_\text{res}$ that maps both the configuration of the objects in the scene and the push proposal direction into a `push centric' frame.}
   \label{fig:push_proposal_image}       
   \end{figure}

\begin{figure}[t]
   	\centering
   	\includegraphics[width=0.90\columnwidth]{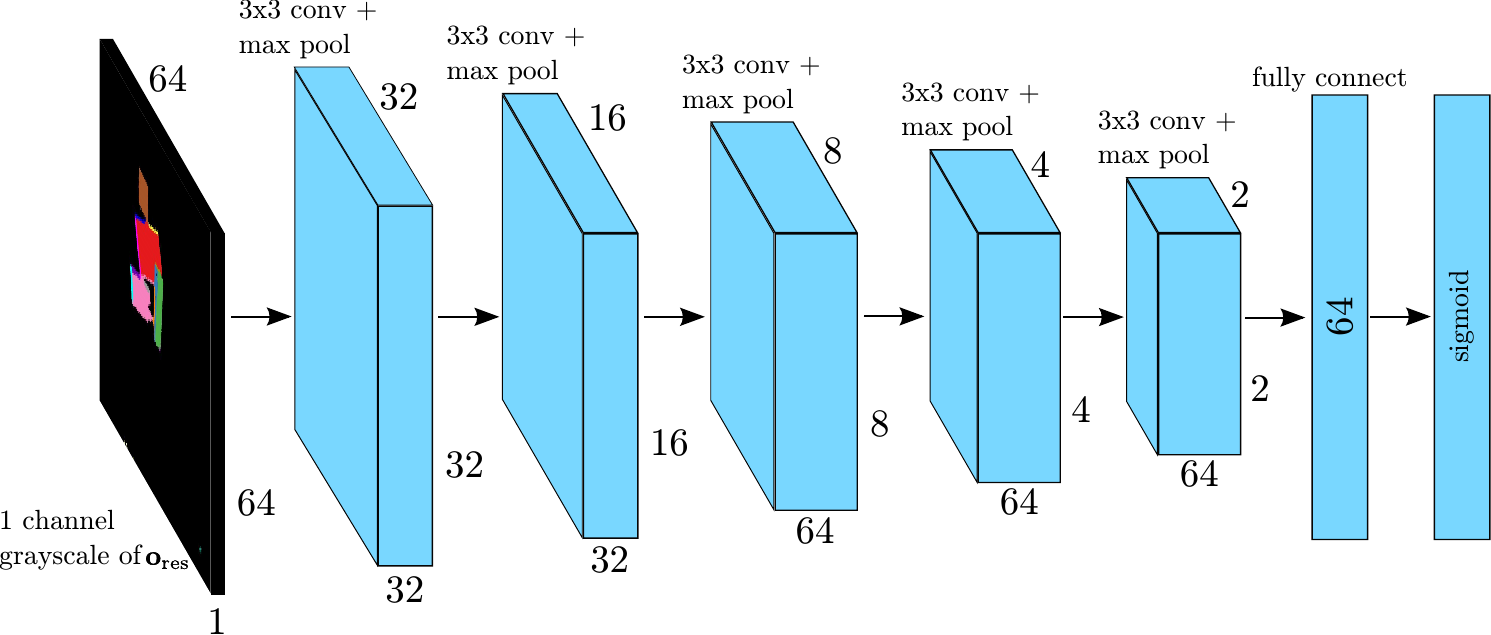}
   	\caption{Our push proposal network architecture to predict a probability of singulation success for a push proposal image $\bo_\text{res}$. The input to the network is a single channel grayscale version of the transformed image $\bo_\text{res}$ with a resolution of $64\times64$ pixels. The network consists of 5 convolutional layers and 2 fully connected layers.}
   	\label{fig:network_architecture}       
   \end{figure}

   We parametrize the predictor $F(\bo,\ba;\btheta)$ using a deep convolutional neural network, denoted as a push proposal CNN in Fig.~\ref{fig:approach}.
   The distinction between $\bo$ and $\ba$ as input is somewhat artificial, because both inputs are combined in a new image $\bo_\text{res}$ which constitutes the final input of the network. To predict the probability of success of a push proposal for the given singulation task the most important feature is the relation of the candidate segment that we want to push and its relation to neighbouring segments. 
   We propose to fuse the segmented camera image with a push proposal using rigid image transformations $\bT_t,\bT_r$, see Fig.~\ref{fig:push_proposal_image}. First, we transform the image by a rigid translation $\bT_t(\bc)$ according to the start position $\bc$ of the push proposal. 
   Second, we apply image rotation $\bT_r(\alpha)$ according to the push angle $\alpha$. 
   The resulting image is $\bo_\text{res}= \bT_r \bT_t \bo$, as depicted in the right image of Fig.~\ref{fig:push_proposal_image}. At test time we forward the set of transformed input images $\{ \bo_\text{res}^1,\dots, \bo_\text{res}^M \}$ into the push proposal network to predict a probability of success for each push proposal in the set $\{ \ba^1,\dots, \ba^M \}$. The network architecture is depicted in Fig.~\ref{fig:network_architecture}. 
   
    We train the network from scratch using random uniform weight initialization and Adam~\cite{kingma2014adam}. We performed network architecture search with 10 different configurations and found that for our task the depicted network architecture yielded best performance on 2 separate validation sets.  
    We experienced a drop in classification performance when increasing the image size, introducing dropout, or removing pooling operations, and did not experience improvement for adding xavier weight initialization or other activation functions besides ReLUs. 
    Finally, we train our model for 25 epochs with a batch size of 64, using the default parameters for Adam (learning rate$=0.001$, $\beta_1=0.9$, $\beta_2=0.999$, $\epsilon=1e-08$, decay$=0.0$).

 
  \subsubsection{Push Motion Planning}
  To find a motion plan for each push proposal we use the LBKPIECE motion planning algorithm provided by an out-of-the-box framework~\cite{sucan2012open}. Our arm motion strategy consists of two steps. In the first step, we plan to reach the target push proposal $\ba^m$. In a second step, given that the gripper reached the desired push start position, the robot performs a straight line push with fixed push length $l_a$.
  We compute motion plans for both steps before execution to avoid that the robot executes the reaching plan and then fails to find a plan for the straight line push. We found our two-step procedure to be more stable than executing a reach and push trajectory at once, due to arm controller errors.

  \section{Experiments}
      \begin{figure}[b]
      \centering
      \includegraphics[width=0.99\columnwidth]{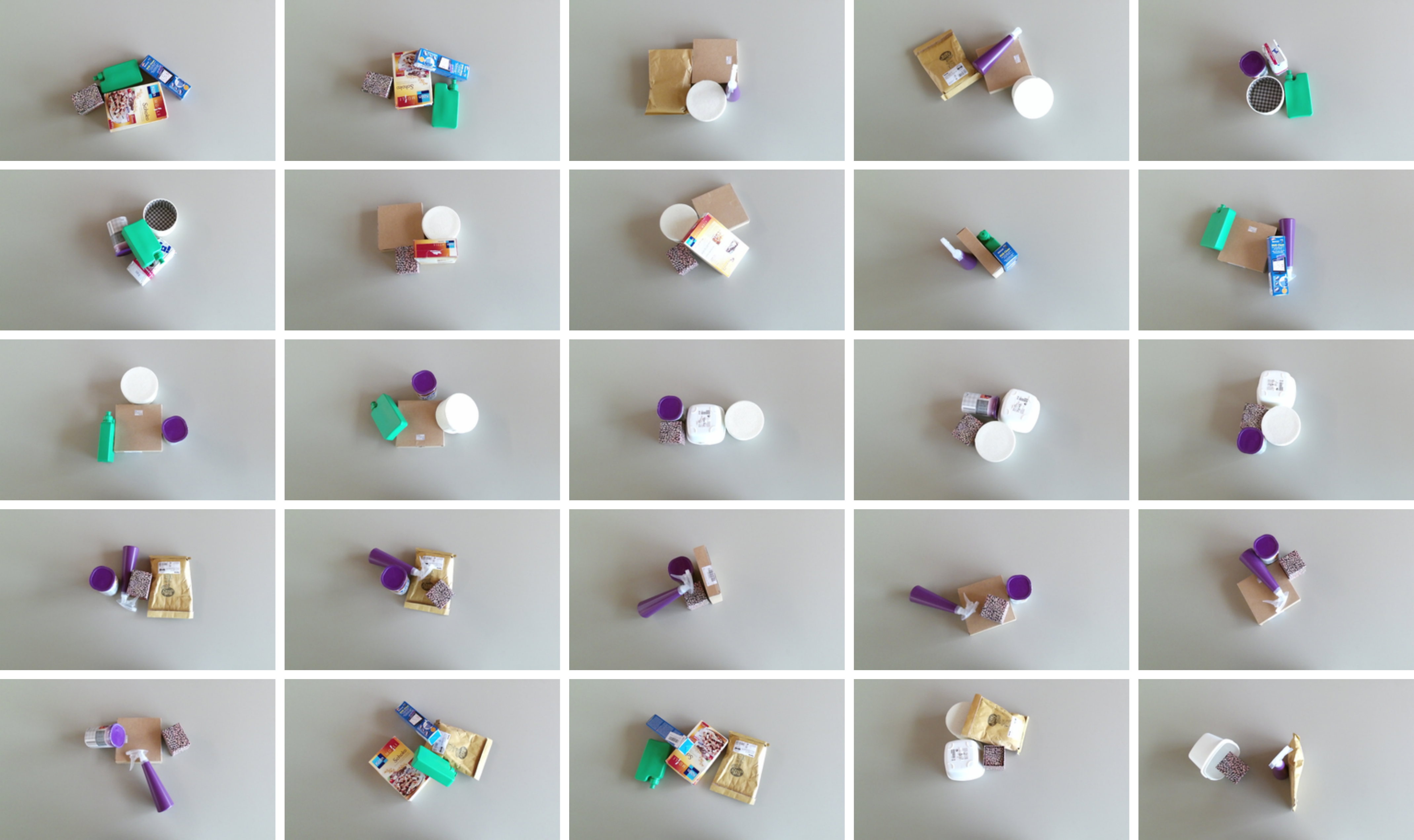}
      \caption{Starting configurations of the 25 test trials with 4 objects.}
      \label{test_scenes}       
      \end{figure}
  
  We conduct both qualitative and quantitative experiments on a real PR2 robot to benchmark the performance of our overall approach. Hereby, we aim to answer the following questions: 1) Can our push proposal network predict meaningful push actions for object shapes and configurations it has not seen during training? 2) Can our network trained in simulation generalize to a real-world scenario? 3) Can our model reason about scene ambiguities including multiple object clusters and focus its attention to objects that are more cluttered than others?  
  In order to answer question (1) we perform real-world experiments with objects that the network has not seen during training and compare against a manual model-free baseline that reasons about available free space in the scene and keeps a history of previous actions. To answer question (2) we test a network that we train solely with data from simulation (denoted as vanilla network). Finally, three challenging experiments with 6 and 8 objects aim to answer question (3), where the learned model has to trade-off between further separating isolated objects to create more space and split isolated object clusters to solve the task.
  The overall intent of our experiments is to show that a learned model is able to directly map from sensor input to a set of possible actions, without prior knowledge about objects, spatial relations or physics simulation.  


  \subsection{Experimental Setup}
  Our PR2 robot is equipped with a Kinect 2 camera mounted on its head that provides RGB-D images with a resolution of $960\times540$.
  We use both arms of the robot, each with 7-DOF to perform the pushing actions. The task is designed as follows: a trial consists of a fixed number of objects from a test set of everyday objects with different shapes that the robot has never seen before. In the starting configuration for a trial all objects are placed or stacked on a table to be in collision. The starting configurations are saved and reproduced for all methods in a manual manner using point cloud alignment. We conduct different experiments of increasing difficulty ranging from 4 to 8 objects, in which for each experiment the number of objects stays fixed. 
  The 25 starting configurations for our quantitative evaluation with 4 objects are depicted in Fig.~\ref{test_scenes}. 
  
  We report the number of successful singulation trials. A singulation trial is considered successful if all objects are separated by a minimum distance of 3cm. Further, we report results for singulation success after every action, to show which methods can successfully execute the task with less actions. For each trial the robot performs a fixed set of actions. Given the number of objects $n_o$ for each experiment the maximum number of pushes is $n_\text{pushes} = \lfloor 1.3\cdot n_o \rfloor + 1$. 
  The robot is allowed to stop before, if it reasons that all objects are singulated. This is implicitly encoded into our method. If the robot does not find a motion plan for the set of push proposals that the network ranked as positive it automatically terminates the trial.  

  To train our push proposal network we use an iterative training procedure. First, the vanilla network is trained on labeled data from a total of 2,486 (243 positives, 2,243 negatives) random interactions performed in simulation. 
  Then the aggregated network is trained on additional 970 interactions (271 positives, 699 negatives), which we collected using the vanilla network in both simulation and real-world, resulting in a training dataset size of 3,456 push interactions. We use the object dataset by Mees~\etal~\cite{mees17iros} for our simulation experiments.

  \subsection{Baseline Method}
  We provide a quantitative comparison against a manually designed baseline method, to evaluate whether or not the predictions of the push proposal network lead to improved object singulation results. We do not provide knowledge about the objects and their physical properties to our model, correspondingly our baseline method is model-free and follows an interactive singulation strategy that is reasonably effective.
  The method which we denote as `free space+tracking' works as follows:
  \begin{enumerate}
  \item{Given a set of segments $\{ \bs^1,\dots, \bs^L \}$ from our object over-segmentation approach we construct a graph that includes all segments, where each segment is a node in the graph and the edges correspond to distances between the segments. We represent each segment as an axis-aligned bounding box (AABB) and compute an edge between 2 segments by means of the Manhattan distance of the 2 AABBs.}  
  \item{We compute 2 features for scoring push proposals with the baseline method. The first feature is a predictive feature that reasons about the resulting free space if a segment would be pushed to some target location.
   To compute the feature, we predict a straight line motion of the respective segment to which the push proposal is assigned to, according to the push direction and the length. 
   Next, we compute the Manhattan distance between the resulting transformed AABB and all other segments in the scene, which we assume will remain static. The free space feature $f_s$ is the minimum distance between the transformed AABB and the other AABBs. If the predicted final position of a segment would lead to collision with another segment the free space feature is zero.}
  \item{The second feature includes the push history that we store for each segment. It follows the intuition that the same segment should not be pushed too often, regardless of the predicted free space around it. To retrieve the push history over all segments, we follow a segment-based tracking approach,
  which aligns the centroid and the principal components of two segments from the current set of segments and the set of segments from the last interaction.
  We match a set of segments using a weighted average of the principal components and the segment centroid distances $d(\bs^l,\bs^m)=0.6\cdot d_{\text{pca}}(\bs^l,\bs^m)+0.4 \cdot d_{c}(\bs^l,\bs^m)$. 
  To punish multiple pushes $r$ of a segment throughout a trial we map the push motion history into a normalized feature using an exponential decay function $f_h=exp(-r)$.}
   \item{Accordingly, a push proposal $\ba^m$ receives the score $p^m=0.5\cdot f_s + 0.5\cdot f_h$.}
  \end{enumerate}

  \subsection{Quantitative Comparisons}

  \begin{figure}[b]
  \centering
  \includegraphics[width=0.70\columnwidth]{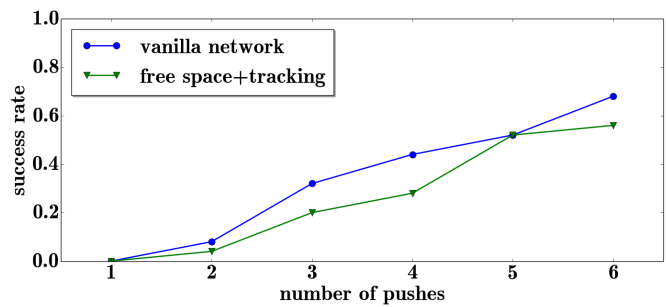}
  \caption{Success rate with respect to number of pushes (max. 6) required to clear 4 objects (experiment 1).}
  \label{fig:4objects_num_pushes}       
  \end{figure}

  We extensively evaluate our method in real-robot object singulation experiments with 4,6 and 8 unknown objects. Additionally we perform an experiment with 6 identical objects to further challenge our approach.
  Our results with 4 objects (experiment 1), shown in Table~\ref{tab:quantitative_results}, indicate that our method is able to improve over the performance of the manual baseline.
  The success rate of our vanilla network is $68\%$, which suggests that the model is making meaningful predictions about which push actions to perform in order to singulate all 4 objects.
  Fig.~\ref{fig:4objects_num_pushes} provides a more fine-grained evaluation, showing the success rate with respect to the number of pushes. Note that the network requires less push actions to complete the task. Interestingly, the baseline method performs on par with the network after the robot has executed five push actions, but then it does not further improve after 6 executed pushes. We noted that the baseline method sometimes fails to singulate the last 2 objects of the scene and instead will choose an object that might already be singulated because it lacks a good attention mechanism.

  \begin{figure}[t]
  \centering
  \includegraphics[width=0.70\columnwidth]{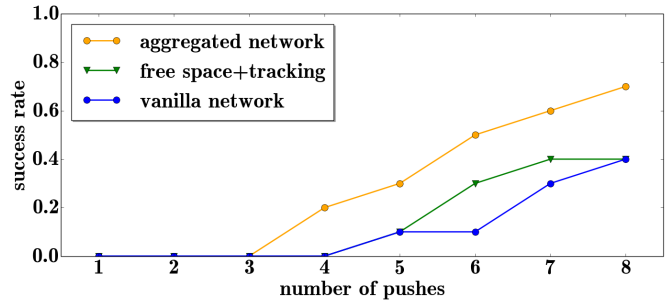}
  \caption{Success rate with respect to number of pushes (max. 8) required to clear 6 objects (experiment 2).}
  \label{fig:6objects_num_pushes}       
  \end{figure}

  \begin{table}[t]
  \caption{Object singulation results for our four experiments of increasing difficulty.}
  \label{tab:quantitative_results}       
  \begin{tabular}{p{1.5cm} p{1.8cm} p{2.8cm}p{1.6cm}p{1.6cm}p{1.8cm}}
  \hline\noalign{\smallskip}
  Experiment No. & Number of objects & Method & Success/total trials & Success rate & Mean number of pushes \\
  \noalign{\smallskip}\svhline\noalign{\smallskip}
  1 & 4 & Free space + tracking & 14/25 & 56\% & 5.4$\pm$1.68\\
   & 4 & Vanilla network & 17/25 & 68\% & 4.96$\pm$1.84 \\
  \noalign{\smallskip}\hline\noalign{\smallskip}  
  2 & 6 & Free space + tracking & 4/10 & 40\% & 7.8$\pm$1.61\\
   & 6 & Vanilla network & 4/10 & 40\% & 8.1$\pm$1.37\\
   & 6 & Aggregated network & 7/10 & 70\% & 6.7$\pm$2.0\\
  \noalign{\smallskip}\hline\noalign{\smallskip} 
  3 & 6 (identical objects) & Aggregated network & 10/20 & 50\% & 7.6$\pm$1.76\\
  \noalign{\smallskip}\hline\noalign{\smallskip}  
  4 & 8 & Aggregated network & 4/10 & 40\% & 11.1$\pm$1.60\\
  \noalign{\smallskip}\hline\noalign{\smallskip}
  \end{tabular}
  \end{table}

  The task is more complex with 6 or 8 objects, due to additional space constraints and formation of multiple object clusters.
  When looking at the scene in Fig.~\ref{fig:approach} one sees that 4 out of the 6 objects are located very closely on the table (green drink bottle, blue box, white round bowl, white coffee bag).
  Although there are many possible push actions that we would consider reasonable, only pushing the coffee bag or the round bowl to the left side of the table can clear the scene.
  Accordingly, we find that the performance of the vanilla network drops with respect to the previous experiment with 4 objects. During training it has only seen a small amount of scenes where a random baseline would have cleared all but 2 objects and even less likely has seen examples where the robot by chance chose the right action to clear the 2 remaining objects. Therefore, in this scenario the vanilla network and the baseline method perform on par with a success rate of $40\%$. When comparing the  average number of pushes the baseline performs slightly better, see Fig.~\ref{fig:6objects_num_pushes}.
  Results show that the aggregated network clearly outperforms the other two methods, winning seven out of ten trials with an average of $6.7$ push actions needed to singulate all 6 objects, as shown in Table~\ref{tab:quantitative_results} (experiment 2). To get an intuition about the numerical results we refer the reader to the performance reported by Hermans~\etal~\cite {hermans2012guided}, who evaluated a very similar task with 6 objects. They report a success rate of $20\%$ with twelve average number of push actions. 
  
  We perform an additional experiment with 6 identical objects (sponges) using the aggregated network and reach a success rate of $50\%$.
  The most challenging experiment is conducted with 8 objects and even though the network was not trained with such large amount of objects on the table our approach performs well and is able to generalize.
  Both experiments are depicted in Table~\ref{tab:quantitative_results} (experiment 3 and 4).

  \subsection{Qualitative Results}
\begin{figure}[t]
  \centering
  \includegraphics[width=0.99\columnwidth]{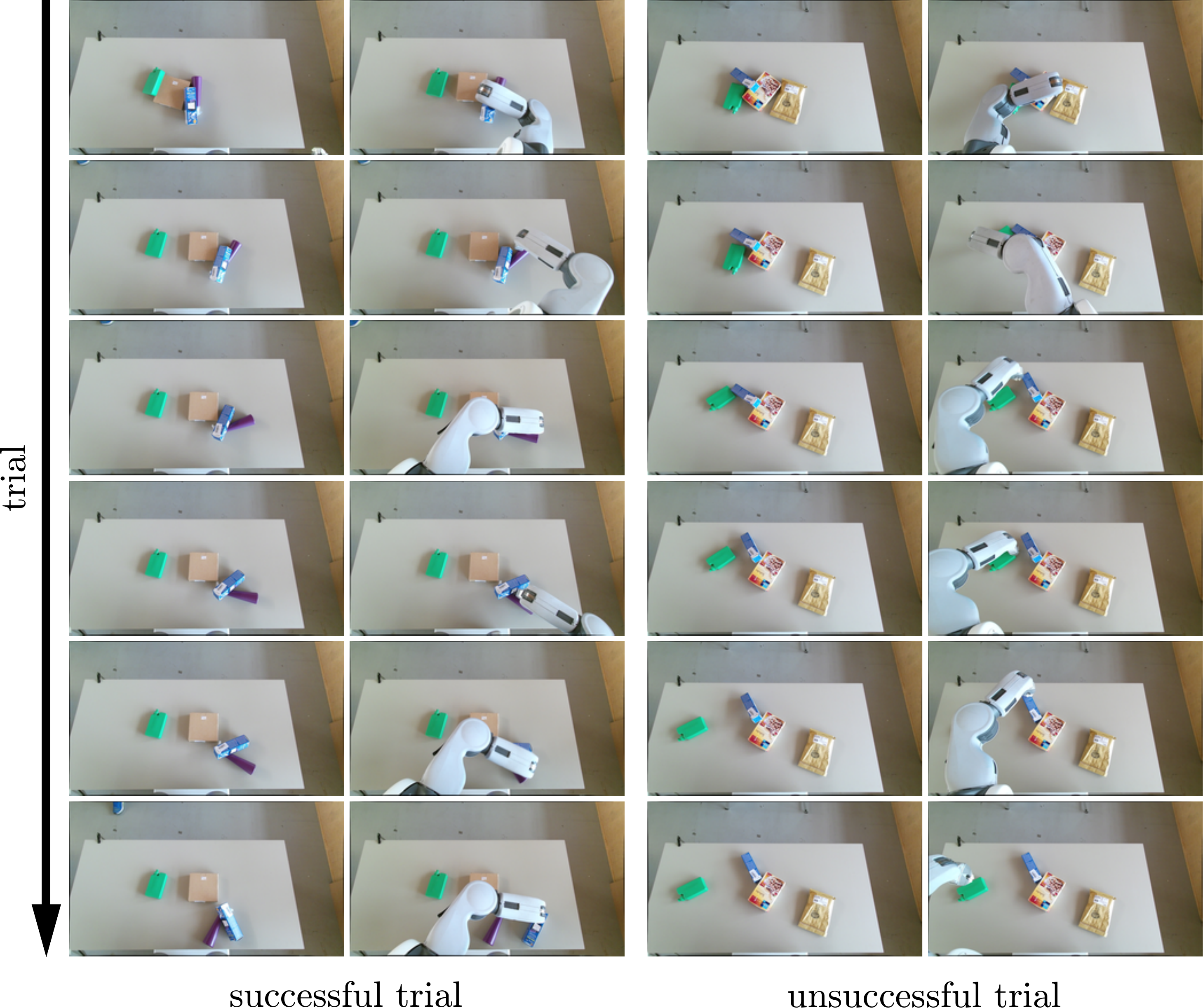}
  \caption{ On the left, we see a successful trial. Note that the robot tries several times to clear a small plastic bottle which is stuck under a blue box and manages to singulate the objects in the very last action. 
  On the right we see a run that fails because of two reasons: First, the robot is not able to perform a save push action (fourth column, fifth row), which results in 2 objects moving more closely together (blue box, cereal box). Second, the network is not able to draw its attention to the 2 objects that are now touching and instead proposes an action that moves the green bottle further away from the scene (lower right image).}
  \label{fig:qualitative_trial}       
  \end{figure}

\begin{figure}[t]
  \centering
  \includegraphics[width=0.90\columnwidth]{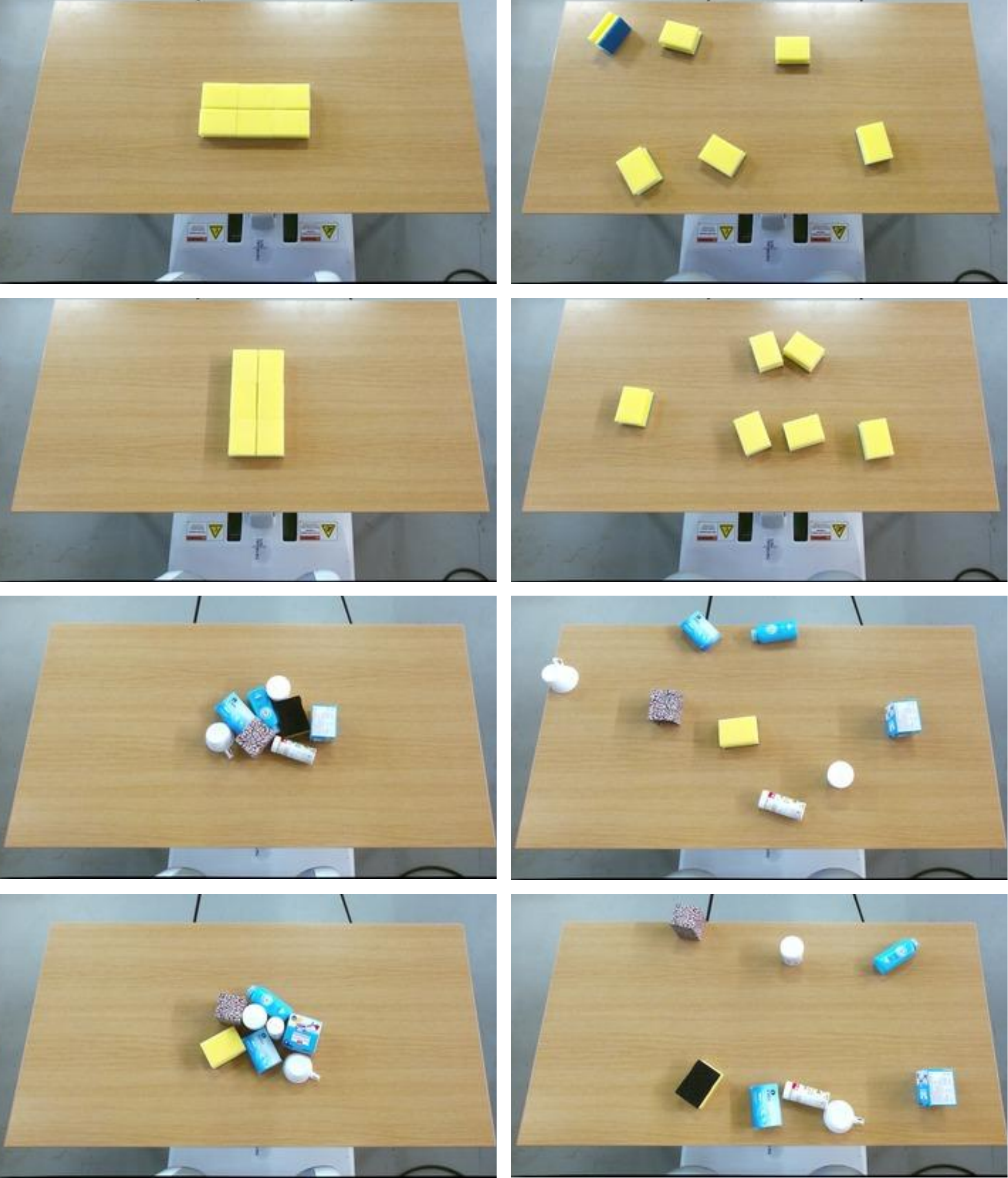}
  \caption{The first row depicts a successful singulation run with six identical objects (experiment 3). The left image depicts the starting configuration and the right image the end result after manipulation. The second row shows a failed trial.
  The third row shows a trial where the robot managed to singulate 8 objects (experiment 4).
  The last row depicts a trial that failed because the object cluster at the bottom is not cleared.}
  \label{fig:qualitative}       
  \end{figure}

  Fig.~\ref{fig:qualitative_trial} gives insights about the singulation trials from the perspective of the Kinect camera for experiment 1. The first and the third column shows the current scene, while the second and fourth column shows the robot performing a push action.
  The trial evolves from top to bottom of the figure.
  Fig.~\ref{fig:qualitative} shows qualitative end results of singulation runs with
  6 identical objects (experiment 3) and 8 objects (experiment 4).

  \section{Conclusions}
  We presented a novel neural network-based approach to singulate unknown objects in clutter by means of pushing actions and showed that it is possible to
  perform challenging perceptual manipulation tasks by learning rich feature 
  representations from visual input. Unlike traditional methods we do not
  manually design features for the task at hand. Instead, we train a high-capacity 
  convolutional neural network that we incorporate into a novel system of motion planning and action selection, leading to good generalization performance, while requiring a reasonable amount of labeled training data.
  We tested our method in extensive real-robot experiments using a PR2 robot and showed the ability of our method to achieve good performance for the challenging task of singulating up to 8 objects. In the future it would be interesting
  to train the network in a self-supervised manner, which poses the challenge
  to automatically generate labels from ambiguous visual data. Furthermore, the 
  network could be extended to a multi-class model that predicts different push lengths, which would require collecting additional training data with varying
  push lengths.
  
  \begin{acknowledgement}
  This work was partially funded by the German Research Foundation under
  the priority program “Autonomous Learning” SPP 1527 and under grant number EXC 108.
  We thank Seongyong Koo for advice with the baseline method.
  We further thank Sudhanshu Mittal, Oier Mees and Tim Welschehold for their help and ideas.
  \end{acknowledgement}

  \bibliographystyle{spmpsci}
  \bibliography{references}

  \end{document}